\PassOptionsToPackage{ruled,vlined,linesnumbered}{algorithm2e}
\documentclass[wcp]{jmlr}
\usepackage{longtable}

\usepackage{booktabs}
\newcommand{\Fview}{\mathcal{F}_{\mathrm{view}}}

\usepackage{lineno}
\SetAlgoLined          
\LinesNumbered          
\makeatother

\makeatletter
\let\Ginclude@graphics\@org@Ginclude@graphics 
\makeatother

\jmlryear{2026}

\title[EVADE]{EVADE: Evidence-Verified Agentic Diagnosis with Escape}

\author{\Name{Mohaimenul Azam Khan Raiaan}\thanks{Equal contribution.} \\
\addr Department of Data Science and Artificial Intelligence, Monash University, Clayton, VIC, 3800, Australia
\AND
\Name{Nur Mohammad Fahad}\footnotemark[1]\\
\addr School of Engineering and Energy, Murdoch University, Murdoch, WA 6150, Australia
}

\begin{document}

\makeatletter
\let \@jmlrpages \@empty
\makeatother

\maketitle

\begin{abstract}
Medical vision-language models (VLMs) can achieve high accuracy, but remain unreliable: they are systematically overconfident, benefit little from test-time reasoning, and lack the ability to reliably calibrate trust in their own responses. We introduce EVADE (Evidence-Verified Agentic Diagnosis with Escape), an inferential, non-training method that enhances the safety of deploying a single frozen VLM. EVADE responds and, when uncertain, localises the region most diagnostically relevant, re-answers on a zoomed view, and commits only when both the entire image and the zoomed view responses agree; otherwise, it abstains. To directly address the verification hallucination in single-model self-checking, our main idea is to verify gate consistency across different image views rather than re-reading the model's own text. Experimental evaluation in VQA-RAD, SLAKE, and PathVQA using Qwen2.5-VL-7B reports that EVADE is the only method that simultaneously improves both calibration and selective risk while maintaining accuracy, educing expected calibration error (ECE) by up to 45\% compared to zero-shot, while chain-of-thought, self-consistency, and self-verification all fail at least one axis. A grounding analysis reports that self-proposed regions perform better at diagnostic structure localisation than centres or random crops. However, a 7B VLM cannot use this localisation to revise answers. Therefore, reliability gains come from the consistency gate and calibrated abstention.
\end{abstract}

\section{Introduction}
Vision-language models (VLMs) are increasingly being proposed as assistants for medical image interpretation, answering free-form questions about radiology and pathology images \citep{vqarad, slake, medgemma}. However, clinical use imposes the requirement that accuracy alone does not capture: a deployable system must signal when its answer is unreliable; thus, uncertain cases are deferred to the clinician rather than acted
upon \citep{atcxr}. Recent evidence indicates that this reliability, not raw accuracy, is the binding constraint for medical VLMs and that standard remedies do not address it.

In VLM-based diagnosis, three failure modes recur more often \citep{dutta2025vision, meddeb2025evaluating}. First, medical VLMs are systematically overconfident and poorly calibrated, and a study spanning three model variants with scales from 2B to 38B parameters finds that this overconfidence is not removed by scaling the model or by prompting strategies such as chain-of-thought or verbalised confidence \citep{byun2026overconfidence}. Second, the remedy of additional test-time reasoning transfers poorly to medical imaging, for instance, chain-of-thought can reduce the accuracy of medical VQA because domain-specific warnings weaken the visual analysis, and the grounding chain causes early perception errors\citep{wu2026better}, and it has also been
shown to induce overconfidence in VLMs \citep{welch2026cost}. Third, the common safety
strategy of asking a model to verify its own answer is unreliable. Because the verifier
and the generator are the same capacity-coupled model, the verifier over-agrees with
the generator and under-attends to the image,  a lazy verifier
produces a verification mirage in which apparent agreement masks a high rate of false
acceptance; in multi-turn loops, this locks in initially wrong answers
\citep{jin2026verification}.

Existing agentic approaches to medical imaging do not close this gap. Tool-augmented
agents such as MedRAX orchestrate external specialist models through a reasoning loop
and achieve strong accuracy, but they depend on a suite of trained tools rather than a
single self-contained VLM, and are not organised around calibration or abstention
\citep{medrax}. Uncertainty-aware triage agents add an abstain-or-defer policy for
chest radiograph classification \citep{atcxr}, and other methods improve VLMs through
training, for example, reinforcement learning with verifiable rewards \citep{huang2026synthesizing}.
The question of how to make a single pretrained VLM reliable purely at inference,
without tools or training, remains underexplored.

We address this question with EVADE (Evidence-Verified Agentic Diagnosis with Escape),
a training-free procedure that runs entirely at inference on a pretrained VLM. EVADE
progresses in three stages. It first answers the full image and estimates its confidence;
if it is confident, it commits. If uncertain, it asks the model to localise the single most
diagnostically relevant region, crops, and upscales that region so that small findings
become legible, and re-answers on this new view. It then commits only when the
whole-image answer and the zoomed-view answer agree with sufficient confidence,
reconciles using both views when they disagree, and abstains when uncertainty remains.

The central design choice is that verification is gated on agreement between two views
of the image, the global view and a self-proposed zoomed view, rather than on the model re-reading its own text. This directly targets the verification mirage: because the second answer is conditioned on fresh visual evidence that the model itself requested,
the verifier cannot authorise the generator without consulting the image, which is
precisely the failure that self-checking is only text \citep{jin2026verification}. The use of a region-of-interest (ROI) view is further motivated by evidence that training-free regions improve medical visual grounding and can reverse chain-of-thought 
degradation \citep{wu2026better}, and abstention prevents the error compounding that
multi-turn self-verification produces.

We make the following contributions.
\begin{itemize}
  \item We propose EVADE, a training-free, inference-only agentic loop that makes a
 single pretrained medical VLM more reliable by coupling an uncertainty-triggered
  evidence zoom, cross-view consistency verification, and calibrated abstention.
  \item We identify cross-view evidence agreements as verification signals that address the verification illusions of single-model self-verification and instantiate them without additional training or external tools.
  \item We design reproducible evaluations of open VLM (Qwen2.5-VL-7B-Instruct) and three public standards (VQA-RAD, SLAKE, PathVQA), which measure reliability, including expected calibration errors, selective prediction risk, and ground overlaps, with full calibration isolation for each component.
  \item EVADE is the only method to improve calibration and selective risk while maintaining accuracy, reducing the expected calibration error by up to $45\%$ compared to zero-shot in three benchmarks of 7B VLM, and the chain of thought, self-consistency, and self-verification fail on at least one axis. An independent analysis also shows that the proposed regions localise diagnostic structures better than centre or random crops. 
\end{itemize}

\section{Related Work}

\subsection{Medical visual question answering}
Medical VQA benchmarks pair clinical images with natural-language questions \citep{lin2023medical, zhang2024development}. VQA-RAD
provides clinician-generated questions on radiology images \citep{vqarad}, SLAKE adds
semantic annotations, including segmentation masks and bounding boxes, together with
bilingual questions \citep{slake}, and PathVQA targets pathology images
\citep{pathvqa}. Strong accuracy in these benchmarks has come largely from models that
are fine-tuned in them, including joint vision-language pretraining and reinforcement
learning with verifiable rewards \citep{huang2026synthesizing}. Medical foundation VLMs such as
MedGemma further improve in-domain performance \citep{medgemma}, while general open VLMs
such as Qwen2.5-VL provide capable zero-shot baselines with native visual grounding
\citep{qwenvl}. Our work differs in setting: we do not fine-tune these datasets and
instead study the inference-time reliability of pretrained models.

\subsection{Agentic and tool-augmented medical reasoning}

A growing number of studies equip medical systems with agent control \citep{wang2025medagent, xia2025mmedagent}. MedRAX integrates specialist chest radiographic tools with multimodal LLMs in an independent logic loop for training and reports state-of-the-art results in complex chest radiographic benchmark \citep{medrax}. AT-CXR introduces an uncertainty-sensitive prioritisation agent that estimates confidence and distribution in each case, develops a step-by-step policy for the issuance, expansion, or delay of decisions, and improves the selective prediction risk in chest radiograph classification \citep{atcxr}. Other systems improve medical VLMs through training, such as data synthesis with a generator-verifier and strengthening learning \citep{huang2026synthesizing}. Our method, EVADE, differs in two ways: it uses a single pre-trained VLM without external tools or training, and its agent's actions are organised around calibration and abstention rather than tool selection or answer maximisation. 

\subsection{Test-time reasoning and its limits in medicine}
Chain-of-thought prompting \citep{wei2022} and self-consistency \citep{wang2023} methods improve reasoning in general settings, and test-time scales usually become a common layer for precision. However, in medical imaging, these methods do not transfer well. Naive token budget scaling can yield limited benefits for most medical VLMs \citep{oh2025rethinking}. Chain-of-thought reduces the accuracy of medical VQA by compounding perceptual uncertainty \citep{wu2026better}, and can cause overconfidence\citep{welch2026cost}. In particular, training-free interference that improves perception, such as indications of region-of-interest, can reverse this degradation \citep{wu2026better}, which motivates us to use evidence-zoom rather than a longer text argument.

\subsection{Calibration, selective prediction, and self-verification}

Confidence calibration is widely used \citep{du2025confidence}, and post-hoc methods, such as temperature scaling, have been shown to reduce calibration errors in deep networks \citep{guo2017}. Selective predictions formalise the option of abstention and are evaluated through risk cover curves and the area below them \citep{geifman2017}. Specifically for medical VLMs, overconfidence persists and is not fixed by prompting, although post-hoc scales help. Self-verification has been proposed as a safety layer, but recent literature has shown that single-model self-verification is unreliable, exhibits a lazy checker effect and verification hallucination, and that cross-checking across image views or models is more reliable \citep{jin2026verification}. The nearest training-free checker, V-Loop, detects hallucinations by visual logic loops and attention coherence \citep{vloop}; it cheques by re-querying with text and checking attention, while EVADE cheques by re-answering on a self-proposed zoomed crop and cross-view agreement gating, coupled with an explicit abstention decision.

\subsection{Gap Analysis}
Previous work \citep{medrax, atcxr, huang2026synthesizing, wei2022, wu2026better, welch2026cost, jin2026verification, vloop} on reliable medical VQA progressed by strengthening models through scaling, pre-training, or synthetic data curation, which increased precision, yet left frozen models over confident and unable to abstain. The addition of inference-time strategy, whether tool-based agents, uncertainty testing, or text-level self-consistency, all control a single view and, therefore, endorsing confident errors. No method evaluates its reliability using new visual evidence collected by the model. EVADE addresses these gaps using an inference loop with no-training establishing a commitment or abstention on the agreement between the whole image and the self-proposed zoom, and turning cross-view consistency into a calibrated selective prediction.
\section{Method}
\label{sec:method}

\subsection{Overview and problem setup}
We study selective medical VQA, in which a model can respond or abstain.
Let $f_\theta$ be a frozen preprained VLM. Given an image
$x\in\mathbb{R}^{H\times W\times 3}$ and a question $q$, autoregressive decoding produces a
token sequence $y=(y_1,\dots,y_T)$ with per-token distributions
$p_\theta(y_t\mid y_{<t},x,q)$, from which we read an answer string $a=\mathrm{parse}(y)$.
We write a single query as $(a,\pi)=f_\theta(x,q)$, where $\pi=\{y,\log p_\theta(\cdot)\}$
carries the decoded answer and its token log-probabilities. A selective predictor is a
pair $(h,g)$ of a classifier $h$ and a binary gate $g$: on input $(x,q)$ it returns
$h(x,q)$ when $g(x,q)=1$ and abstains, denoted $\bot$, otherwise \citep{geifman2017}.
Its quality is summarised by the risk-coverage trade-off, where coverage is the fraction of answered cases, and selective risk is the error rate of the selected cases.

\begin{figure}[ht!]
\includegraphics[scale=0.5]{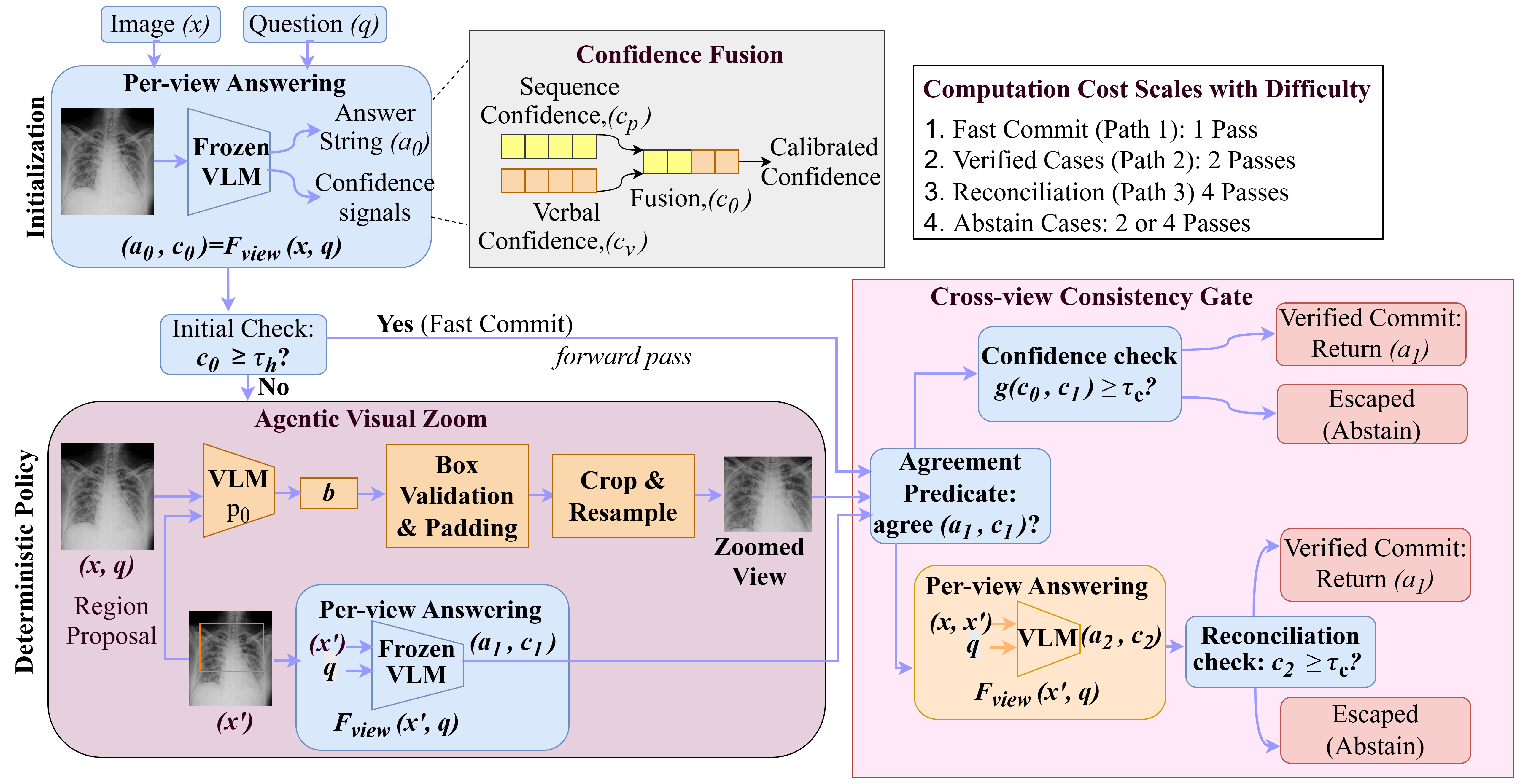}
\caption{Proposed methodology diagram of EVADE, showing the three steps, with model, and conditions} 
\label{fig:method}
\end{figure}

EVADE manifested $(h,g)$ as a single deterministic policy on $f_\theta$ that
requires no fine-tuning and no auxiliary networks. Three designs follow from the failure modes in Section 1: the confidence used to gate must not rely on a single
overconfident signal; verification must consult visual evidence rather than re-read
text; and the system must be able to abstain rather than commit a conflicted answer.
EVADE meets these through (i) a calibrated first answer, (ii) an agentic visual
zoom that re-observes a self-selected region, and (iii) a cross-view consistency gate that commits, reconciles, or abstains. The Algorithm~\ref{alg:evade} gives the full
procedure. In addition, Figure \ref{fig:method} presents the methodology diagram of EVADE, highlighting the execution flow and conditions.

\subsection{Per-view answering and confidence}
\label{sec:conf}
Each stage queries $f_\theta$ on one view, resulting in an answer and a
scalar confidence $c\in[0,1]$. We combine two complementary signals. The \textbf{sequence confidence} is the geometric mean of the answer-token probabilities,
\begin{equation}
  c_p \;=\; \exp\!\Big( \tfrac{1}{|\mathcal{A}|}\!\sum_{t\in\mathcal{A}}
    \log p_\theta\big(y_t \mid y_{<t}, x, q\big) \Big),
  \label{eq:cp}
\end{equation}
In Eq. \eqref{eq:cp}, $\mathcal{A}$ indexes the answer span tokens; $c_p$ is the internal 
probability of the model and is sensitive to verbal uncertainty. Verbalised \textbf{ confidence}
$c_v\in[0,1]$ is evoked by instructing the model to add a confidence value to its
answer, capturing self-assessment that probability alone could miss. However, verbalised confidence can itself be systematically overconfident in medical VLMs \citep{senoglu2026just}, therefore, it is not trustworthy in isolation, but the fusion of the two signals expressed in Eq. \eqref{eq:fuse},
\begin{equation}
  c \;=\; \lambda\, c_v + (1-\lambda)\, c_p, \qquad \lambda\in[0,1],
  \label{eq:fuse}
\end{equation}
with $\lambda$ selected from the delayed (development) data. (Section~\ref{sec:setup})We denote the
per-view outputs $(a,c)=\Fview(x,q)$, folding Eqs.~\eqref{eq:cp}--\eqref{eq:fuse} into
$\Fview$.

\subsection{Agentic visual zoom}
\label{sec:zoom}
The first answer is computed on the full image $(a_0,c_0)=\Fview(x,q)$. When
$c_0\!\ge\!\tau_h$, the case is easy, then EVADE commits immediately; otherwise, it goes to the next step, and takes a perceptual action. It asks the model to localise the most diagnostically relevant region for the question, using VLM's native grounding ability, obtained using Eq.\eqref{vlm}
\begin{equation}
  \tilde{b} = \rho_\theta(x,q), \qquad \tilde{b}=(u_1,v_1,u_2,v_2).
  \label{vlm}
\end{equation}
Then, the bounding box is validated and projected onto the image domain, returning to a fixed central window $b_c$ that covers a fraction $\gamma$ of each side when the output is degenerate using Eq. \eqref{eq:box},
\begin{equation}
  b =
  \begin{cases}
    \Pi_{[0,W]\times[0,H]}(\tilde{b}) & \text{if } \tilde{b} \text{ is valid and }
      \mathrm{area}(\tilde{b}) \ge a_{\min},\\[3pt]
    b_c & \text{otherwise.}
  \end{cases}
  \label{eq:box}
\end{equation}
By dilating the box by a $p$ margin to maintain the surrounding lesional context, and by upsampling, we resample crops to a fixed central window of $S$,
\begin{equation}
  x' \;=\; \mathcal{R}_S\!\big(\mathrm{crop}(x,\,\mathrm{pad}(b,p))\big),
  \qquad S \ge \max(H,W),
  \label{eq:crop}
\end{equation}

After cropping in Eq. \eqref{eq:crop}, the area of interest occupies more input tokens and fine structures beyond the model's effective resolution threshold. This is the mechanism by which EVADE counteracts bottlenecks in medical perception that restrict chain-of-thought reasoning \citep{wu2026better}. Instead of adding text reasoning on the same pixel, it increases visual evidence for decisions. The second answer is obtained on a zoom view, $(a_1,c_1)=\Fview(x',q)$. By construction, $a_1$ is based on another observation than $a_0$, so the next gate becomes a test of evidence, not of text. 

\subsection{Cross-view consistency as an evidence-grounded verifier}
\label{sec:gate}
\noindent\textbf{Agreement predicate.}\;
Let $n(\cdot)$ normalise an answer by punctuation and article removal, casefolding, and canonicalisation of polarity terms. For closed questions, the agreement is exact,
$\mathrm{agree}(a_0,a_1)=\mathbf{1}\!\left[n(a_0)=n(a_1)\right]$; for open questions, we threshold the token-overlap $F_1$,
$\mathrm{agree}(a_0,a_1)=\mathbf{1}\!\left[F_1\!\big(n(a_0),n(a_1)\big)\ge\theta\right]$.

\noindent\textbf{Why cross-view rather than self-rereading }\;
Naive self-verification re-queries $f_\theta$ in the same $(x,q)$. The checked answer and the original are both samples of $p_\theta(\cdot\mid x,q)$, sharing the same input and model, and tend to agree even when wrong. This is called agreement bias, and the lazy-verification effect produces verification illusions \citep{jin2026verification}. It is conditional on identical images and does not provide new proof of correctness. EVADE, in contrast, measures the stability of the answers under controlled visual intervention 
$x'=\mathcal{R}_S(\mathrm{crop}(x,\rho_\theta(x,q)))$, a self-generated counterfactual view that zooms in on the model's ROI and discards the surrounding context. Agreement over views demonstrates that the prediction is invariant to changes in the model's evidence as a whole, a property that cannot be verified by resampling the text on the fixed image. Subsequently, disagreements are diagnostic; they isolate answers that depend on low-resolution shortcuts or global context rather than the diagnostic region, and EVADE routes them to abstention. In fact, the verification is re-modelled from the text self-consistency problem, which requires the collection of reasoning chains from an image, to the view-consistency problem, which requires the collection of visual views, which is the natural axis of change for a perception-bound task.

\noindent\textbf{Decision policy.}\;
The aggregated confidence of two agreeing views is $g(c_0,c_1)=\max(c_0,c_1)$: EVADE commits the agreed answer when the most confident of the two views clears $\tau_c$. We used the maximum in all reported experiments; the arithmetic and harmonic means are more conservative alternatives. In commitment, the stored confidence is the same $\max(c_0,c_1)$, so the calibration and selective-risk metrics are calculated against it. The policy is denoted in 
Eq.~\eqref{eq:policy}:
\begin{equation}
  \mathrm{EVADE}(x,q) =
  \begin{cases}
    a_0, & c_0 \ge \tau_h, \\[3pt]
    a_1, & c_0 < \tau_h,\; \mathrm{agree}(a_0,a_1),\; g(c_0,c_1) \ge \tau_c, \\[3pt]
    a_2, & c_0 < \tau_h,\; \neg\,\mathrm{agree}(a_0,a_1),\; c_2 \ge \tau_c, \\[3pt]
    \bot, & \text{otherwise,}
  \end{cases}
  \label{eq:policy}
\end{equation}
where, in disagreement, $(a_2,c_2)=\Fview(\{x,x'\},q)$, there is a reconciliation pass that takes into account both points of view to enable the model to settle the issue using all the information at hand. The abstention branch, which prevents conflicting cases from accumulating into confident errors, is reached when opinions only marginally agree or disagree without a confident resolve \citep{jin2026verification}.

\subsection{Selective behaviour and operating point}
Eq.~\eqref{eq:policy} expresses a selective predictor whose coverage is controlled by
the threshold pair $\tau=(\tau_h,\tau_c)$. Increasing $\tau_h$ sends more cases through
verification, while increasing $\tau_c$ tightens the level of commitment and increases
abstention. Sweeping $\tau$ traces an operational curve in the risk-coverage space. We choose a single $\tau$, as well as the fusion weight $\lambda$, by shrinking the area under the risk-coverage curve of the development-set. The test distribution is not used for tuning.
EVADE differs from post-hoc calibration, which rescales confidences but maintains the decision boundary. Inject newly generated visual evidence before making the decision, affecting the response to which is returned.

\subsection{Computational cost}
EVADE uses computation only when necessary. A scenario that clears $\tau_h$ makes a single forward pass. In an ambiguous instance, only four options are used: the initial answer, the region proposal, the zoomed answer, and reconciliation if there is disagreement. The cost increases with the difficulty of the case rather than evenly. The fast-commit proportion is controlled by $\tau_h$.
\begin{algorithm2e}[t]
\caption{EVADE: evidence-verified agentic diagnosis with escape}
\label{alg:evade}
\KwIn{frozen VLM $f_\theta$, image $x$, question $q$; thresholds $\tau_h,\tau_c$; fusion weight $\lambda$}
$(a_0,c_0) \gets f_\theta(x,q)$ \tcp*{answer full image; Eqs.~\eqref{eq:cp}--\eqref{eq:fuse}}
\If{$c_0 \ge \tau_h$}{
  \Return $a_0$ \tcp*{fast commit, one forward pass}
}
$b \gets \rho_\theta(x,q)$ \tcp*{self-proposed region; Eq.~\eqref{eq:box}}
$x' \gets \mathcal{R}_S(\mathrm{crop}(x,\mathrm{pad}(b,p)))$ \tcp*{zoom to fresh pixels; Eq.~\eqref{eq:crop}}
$(a_1,c_1) \gets f_\theta(x',q)$\;
\If{$\mathrm{agree}(a_0,a_1)$}{
  \If{$g(c_0,c_1) \ge \tau_c$}{
    \Return $a_1$ \tcp*{verified commit}
  }
  \Else{
    \Return $\bot$ \tcp*{weak agreement: abstain}
  }
}
\Else{
  $(a_2,c_2) \gets f_\theta(\{x,x'\},q)$ \tcp*{reconcile on both views}
  \If{$c_2 \ge \tau_c$}{
    \Return $a_2$\;
  }
  \Else{
    \Return $\bot$\;
  }
}
\end{algorithm2e}
\subsection{Component ablations}
To understand EVADE's behaviour, we examine lesion policies that remove a single mechanism from Eq.~\eqref{eq:policy}. \textbf{(a) No-zoom} replaces the zoomed view with another read of the full image, reducing the gate to text-based self-verification. \textbf{(b) No-gate} commits the zoomed response when the model zooms, eliminating the consistency test. \textbf{(c) No-abstain} requires a commit in all branches, eliminating the escape. \textbf{(d) Fixed-region} substitutes $\rho_\theta$ with a centre or random crop, eliminating agentic region selection. The contrast in (d) with the whole model isolates the value of where the model looks, while (a) isolates the value of looking at fresh pixels.

\subsection{Experimental setup}
\label{sec:setup}

\paragraph{Datasets.} Three public medical VQA benchmarks are used for our evaluation: PathVQA \citep{pathvqa}, SLAKE (English) \citep{slake}, and VQA-RAD \citep{vqarad}. We measure grounding overlap using the region annotations that SLAKE additionally provides. Both results of both closed-ended and open-ended questions are reported in the following sections.

\paragraph{Model.} We assess using Qwen2.5-VL-7B-Instruct \citep{qwenvl}, a standard open VLM that is used without any fine-tuning and provides token log-probabilities and native visual grounding. Any VLM that can report a region and confidence can use EVADE, which is model-agnostic. Future research will expand the study to include more and larger VLMs.

\paragraph{Baselines.} We compare against zero-shot direct answering, chain-of-thought
prompting \citep{wei2022}, self-consistency \citep{wang2023}, naive self-verification
that asks the same image, and confidence-threshold abstention.

\paragraph{Metrics.} Beyond the accuracy of the answers to the questions, we measure reliability. We report
expected calibration error (ECE) \citep{guo2017}, calculated by dividing predictions by
confidence and averaging the gap between accuracy and confidence. We report
selective-prediction performance through the risk-coverage curve and the area under it
(AURC) \citep{geifman2017}, together with risk at fixed coverage. On SLAKE we measure grounding overlap as the intersection-over-union
between the self-proposed region and the annotated region.

\paragraph{Implementation details.} Committal answers use greedy decoding, while
self-consistency uses temperature sampling. The thresholds $\tau_h$ and $\tau_c$ and the
fusion weight $\lambda$ are selected by grid search in a holdout development split,
minimising development-set AURC; no test data is used for tuning. Statistical significance
is assessed with paired bootstrap $95\%$ confidence intervals and McNemar tests ~\citep{pembury2020effective} across
datasets. All runs use a single NVIDIA GB10 accelerator (DGX Spark); we report the mean
number of forward passes per item (Sec.~\ref {sec:res_cost}) as a hardware-independent cost
measure rather than wall-clock time.


\begin{figure}[ht!]
\centering
\includegraphics[scale=.9]{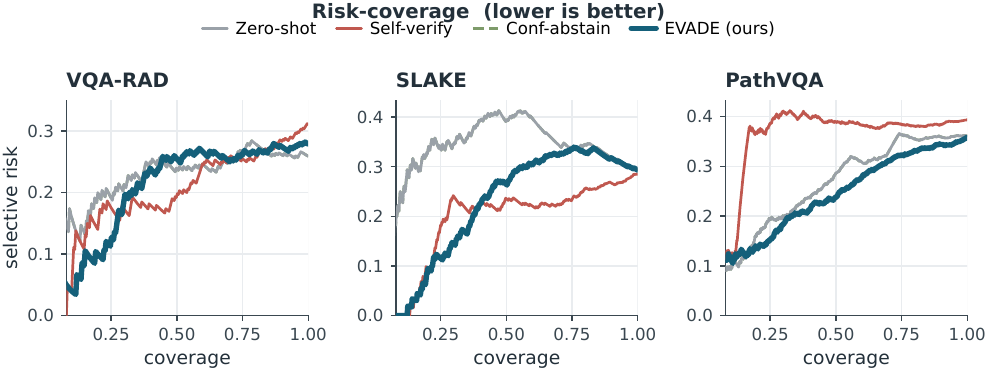}
\caption{Risk-coverage curves on the three benchmarks (Qwen2.5-VL-7B); lower is better.
EVADE attains the lowest selective risk at high coverage. On SLAKE, confidence-thresholded
abstention (Conf-abstain) degrades below zero-shot because raw confidence is anti-correlated
with correctness there; EVADE's cross-view gate recovers a usable selective signal.}
\label{fig:rc}
\end{figure}

\begin{figure}[ht!]
\centering
\includegraphics[scale=1]{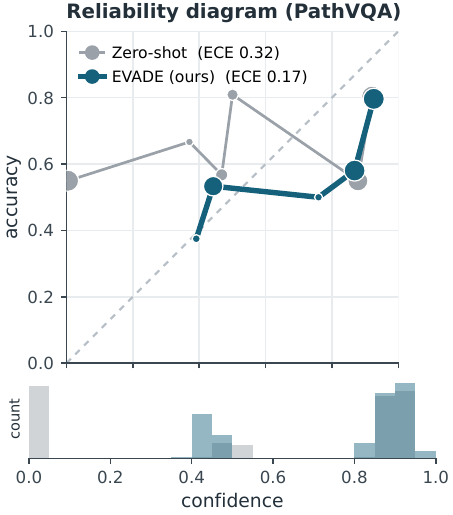}
\caption{Reliability diagram on PathVQA ($10$ equal-width confidence bins; marker area is
proportional to bin population, and the lower panel shows the confidence histogram).
Zero-shot is severely overconfident (ECE $0.32$); EVADE tracks the diagonal far more
closely (ECE $0.17$).}
\label{fig:reliability}
\end{figure}

\begin{figure}[ht!]
\centering
\includegraphics[scale=1]{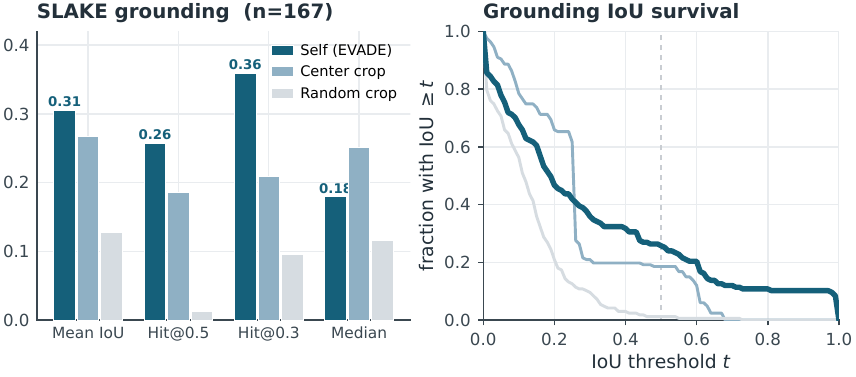}
\caption{SLAKE grounding ($n{=}167$ zoomed items with a region annotation). \textbf{Left:}
self-proposed regions achieve the highest mean IoU and tight-overlap hit-rates; centre
crop attains a higher median but far fewer tight hits. \textbf{Right:} IoU survival; EVADE
matches center crop at loose thresholds and dominates for $\mathrm{IoU}\!\ge\!0.4$,
indicating it more often localises tightly. No self-proposal falls back to the centre
window.}
\label{fig:grounding}
\end{figure}
\newcommand{\gr}[1]{\textcolor{gray!75}{#1}}   

\section{Results and Analysis}
\label{sec:results}

Experimental evaluation is conducted on Qwen2.5-VL-7B-Instruct across VQA-RAD, SLAKE, and PathVQA datasets. All decision thresholds are tuned in the development split (Sec.~\ref{sec:setup}); however, the test sets are not tuned. For each method, we compute selective accuracy (accuracy in the answers), coverage, expected calibration error (ECE), the area under the risk-coverage curve (AURC), and the average number of forward passes per item. Moreover, statistical significance is assessed with paired bootstrap 95\% confidence intervals and McNemar tests across datasets.

Our main finding is that reliability is a combination of three requirements, improved selective risk, improved calibration, and preserved accuracy, and EVADE is the lone method that satisfies all three simultaneously. Each baseline
fails at least one axis; for example, chain-of-thought and self-consistency collapse in accuracy, self-verification fails in calibration, and confidence-threshold abstention can be inversely correlated with accuracy.

\begin{table*}[t]
\centering
\caption{Main results on Qwen2.5-VL-7B. Acc is selective accuracy (on answered items)
at the listed coverage (Cov). \textbf{Bold}: best among the reliability-oriented methods
(zero-shot, self-verify, conf-abstain, EVADE). CoT and self-consistency (\gr{gray})
attain low ECE only by collapsing accuracy, and are shown for reference. EVADE is the
only method that lowers ECE on all three datasets while preserving accuracy;
self-verify lowers AURC but at a large calibration cost.}
\scriptsize
\begin{tabular}{l cccc cccc cccc}
\toprule
& \multicolumn{4}{c}{\textbf{VQA-RAD}} & \multicolumn{4}{c}{\textbf{SLAKE}} & \multicolumn{4}{c}{\textbf{PathVQA}} \\
\cmidrule(lr){2-5}\cmidrule(lr){6-9}\cmidrule(lr){10-13}
Method & ECE$\downarrow$ & AURC$\downarrow$ & Acc$\uparrow$ & Cov
       & ECE$\downarrow$ & AURC$\downarrow$ & Acc$\uparrow$ & Cov
       & ECE$\downarrow$ & AURC$\downarrow$ & Acc$\uparrow$ & Cov \\
\midrule
Zero-shot          & 0.241 & 0.225 & 0.741 & 1.00 & 0.393 & 0.327 & 0.707 & 1.00 & 0.317 & 0.255 & 0.639 & 1.00 \\
\gr{CoT}           & \gr{0.196} & \gr{0.400} & \gr{0.554} & \gr{1.00} & \gr{0.189} & \gr{0.396} & \gr{0.543} & \gr{1.00} & \gr{0.155} & \gr{0.554} & \gr{0.302} & \gr{1.00} \\
\gr{Self-consist.} & \gr{0.149} & \gr{0.311} & \gr{0.562} & \gr{1.00} & \gr{0.164} & \gr{0.246} & \gr{0.560} & \gr{1.00} & \gr{0.300} & \gr{0.595} & \gr{0.317} & \gr{1.00} \\
Self-verify        & 0.519 & \textbf{0.195} & 0.689 & 1.00 & 0.456 & \textbf{0.187} & 0.714 & 1.00 & 0.493 & 0.352 & 0.606 & 1.00 \\
Conf-abstain       & 0.241 & 0.225 & 0.716 & 0.785 & 0.393 & 0.327 & 0.596 & 0.464 & 0.317 & 0.255 & 0.680 & 0.557 \\
\textbf{EVADE (ours)} & \textbf{0.184} & 0.202 & 0.720 & 0.968 & \textbf{0.282} & 0.216 & 0.669 & 0.849 & \textbf{0.173} & \textbf{0.242} & 0.689 & 0.705 \\
\bottomrule
\end{tabular}
\label{tab:main}
\end{table*}

\subsection{Calibration and selective prediction}
\label{sec:res_calib}
Table~\ref{tab:main} presents that EVADE reduces ECE in zero-shot testing on all datasets, by
$23\%$ in VQA-RAD ($0.241\!\to\!0.184$), $28\%$ in SLAKE ($0.393\!\to\!0.282$) and
$45\%$ in PathVQA ($0.317\!\to\!0.173$), and consistently achieves lower AURC on every dataset
(Fig.~\ref{fig:rc}). In addition, these calibration gains are statistically significant as the paired bootstrap $95\%$ CI on $\Delta$ECE against zero-shot in all three datasets are ($[-0.10,-0.01]$ in VQA-RAD, $[-0.17,-0.05]$ in SLAKE, $[-0.17,-0.12]$ in PathVQA), and McNemar tests show no substantial changes in accuracy on any dataset, therefore, the improved calibration has no quantifiable accuracy cost. The reduction in AURC in zero-shot setting is ($95\%$ CI $[-0.15,-0.08]$) in SLAKE and directionally consistent; however, it is not statistically significant in VQA-RAD and PathVQA. The reliability diagram (Fig. (\ref{fig:reliability}) shows that zero-shot is considerably overconfident, accurate well below its declared confidence, both in the mid-range and above it at the top, while EVADE tracks the diagonal much more closely. Both chain-of-thought and self-consistency methods appear to be well-calibrated
in isolation; however, due to their frequent uncertainty, overall
accuracy falls to $0.30$ and $0.32$ on PathVQA from a zero-shot $0.64$, and the corresponding AURC is also poor. In contrast, EVADE improves the calibration while preserving the accuracy.

\subsection{Self-verification and confidence abstention are insufficient}
\label{sec:res_baselines}
Both reliability baselines fail in instructive, contradictory ways. Self-verification improves answer rankings by achieving the lowest AURC on VQA-RAD and SLAKE, but its calibration fails, and ECE increases to $0.52$, $0.46$, and $0.49$. The distance to EVADE in $\Delta$ECE is large and significant across all datasets (up to $-0.34$, all $95\%$ CIs excluding zero), and on the PathVQA dataset, EVADE is remarkably more accurate (McNemar $p<0.001$). This is the sign of verification-mirage, a confident verifier that governs the generator without re-examining the image \citep{jin2026verification}. Besides, confidence-thresholded abstention fails disparately where selective accuracy ($0.596$) drops below the
full-coverage accuracy ($0.707$) at coverage $0.46$, hence, raw confidence is
inversely correlated with accuracy and abstaining discards accurate answers, and the rising Conf-abstain curve in Fig.~\ref{fig:rc} (SLAKE), clearly indicates this. The Cross-view consistency gate of EVADE avoids both pathologies and produces a useful, selectively calibrated confidence signal.

\begin{table}[t]
\centering
\caption{SLAKE grounding ($n{=}167$ zoomed items with a region annotation). Self-proposed
regions achieve the highest mean IoU and tight-overlap hit-rates; no self-proposal fell
back to the centre window ($0\%$). Centre crop has a higher median but far fewer
tight hits, reflecting reliable moderate overlap without precise localisation.}
\scriptsize
\begin{tabular}{l cc ccc cc}
\toprule
Region & Mean & Med. & Hit & Hit & Hit & IoU$\mid$ & IoU$\mid$ \\
       & IoU  & IoU  & @.5 & @.3 & @.1 & cor. & wr. \\
\midrule
\textbf{Self (EVADE)} & \textbf{0.306} & 0.180 & \textbf{0.257} & \textbf{0.359} & 0.677 & 0.319 & 0.277 \\
Center crop           & 0.267 & \textbf{0.250} & 0.186 & 0.210 & \textbf{0.784} & 0.259 & 0.288 \\
Random crop           & 0.128 & 0.115 & 0.012 & 0.096 & 0.563 & 0.129 & 0.124 \\
\bottomrule
\end{tabular}
\label{tab:grounding}
\end{table}

\subsection{Does the agentic zoom localise?}
\label{sec:res_grounding}
In the $167$ SLAKE samples that triggered the zoom and executed a region annotation, the self-proposed regions achieve a mean (mIoU) of $0.306$, which is above the centre crops ($0.267$) and far above the random crops ($0.128$), and it can be mathematically expressed as self $>$ centre $>$ random holds throughout
(Table~\ref{tab:grounding}, Fig.~\ref{fig:grounding}). The distinction broadens at hard
thresholds where EVADE reaches $\mathrm{IoU}\!\ge\!0.5$ on $25.7\%$ of items versus $1.2\%$
at random, with a $21\times$ margin, and the IoU curve displays that EVADE dominates everywhere and outperforms the centre crop beyond $\mathrm{IoU}\!=\!0.4$. The centre crop has a comparatively
higher median IoU ($0.250$ vs.\ $0.180$) and fewer tight hits, securing moderate overlap without precise localisation, while our self-proposal EVADE more often positions
exactly on the diagnostic region. Moreover, no self-proposal fell back to the centre window, and
the IoU region is marginally higher on correctly answered items ($0.319$ vs.\ $0.277$). Consequently, the model
localises the relevant region without prior supervision.

\begin{table*}[t]
\centering
\caption{Component ablations on Qwen2.5-VL-7B. Removing the gate degrades ECE and AURC on
VQA-RAD and SLAKE, identifying it as the main source of the reliability gains. Removing
abstention leaves ECE/AURC unchanged (these are computed over the full confidence ranking)
but forces coverage to $1.0$. Removing the zoom or swapping the self-proposed region for a
centre or random crop barely moves the downstream metrics, despite the large grounding
differences in Table~\ref{tab:grounding}.}
\scriptsize
\begin{tabular}{l cc cc cc c}
\toprule
& \multicolumn{2}{c}{\textbf{VQA-RAD}} & \multicolumn{2}{c}{\textbf{SLAKE}} & \multicolumn{2}{c}{\textbf{PathVQA}} & \\
\cmidrule(lr){2-3}\cmidrule(lr){4-5}\cmidrule(lr){6-7}
Variant & ECE$\downarrow$ & AURC$\downarrow$ & ECE$\downarrow$ & AURC$\downarrow$ & ECE$\downarrow$ & AURC$\downarrow$ & Calls \\
\midrule
\textbf{EVADE (full)}      & 0.184 & 0.202 & 0.282 & 0.216 & 0.173 & 0.242 & 2.10--2.39 \\
\;\; $-$ zoom (text re-read)& 0.186 & 0.206 & 0.290 & 0.215 & 0.182 & 0.249 & 1.50--1.77 \\
\;\; $-$ gate              & 0.197 & 0.218 & 0.320 & 0.237 & 0.168 & 0.242 & 1.87--2.32 \\
\;\; $-$ abstain           & 0.184 & 0.202 & 0.282 & 0.216 & 0.173 & 0.242 & 2.10--2.39 \\
\;\; center crop           & 0.184 & 0.198 & 0.279 & 0.205 & 0.188 & 0.247 & 1.51--1.76 \\
\;\; random crop           & 0.187 & 0.199 & 0.275 & 0.207 & 0.188 & 0.244 & 1.53--1.78 \\
\bottomrule
\end{tabular}

\label{tab:ablation}
\end{table*}

\subsection{Ablations: where the gains originate}
\label{sec:res_ablation}
Table~\ref{tab:ablation} outlines the contribution of each sub-component. The worst performance is observed after removing the
consistency gate for both ECE and AURC in SLAKE, and VQA-RAD (SLAKE ECE
$0.282\!\to\!0.320$, AURC $0.216\!\to\!0.237$), determining the gate as an influential
driver of the reliability gains. On the other hand, removing abstention keeps ECE and AURC unchanged, and was quite
expected since they are computed over the full confidence ranking, but forces coverage to
$1.0$ and removes the selective-accuracy advantages, confirming that abstention drives the
operating point rather than the calibration. However, the most critical part is that removing the zoom or replacing
the self-proposed region with a random or centre crop hardly changes the metrics (SLAKE ECE $0.282$, $0.279$, $0.275$ for self, centre and random), despite their grounding quality differs sharply (Table~\ref{tab:grounding}). Together with
Sec.~\ref{sec:res_grounding}, this outlines our key analysis result that the agentic zoom
precisely localises the diagnostic region, yet a 7B VLM does not interpret that
localisation into changed answers; therefore, the measured reliability gains arise from the 
consistency gate and calibrated abstention instead of the zoomed evidence itself.

\subsection{Efficiency}
\label{sec:res_cost}
EVADE computes on hard cases where the zoom triggers on $44\%$ of SLAKE items ($181/416$) while the rest of the samples are committed in a single pass. In addition, mean forward passes per item are
$2.10$, $1.98$, and $2.39$ across three datasets, against $5.0$ for self-consistency, making EVADE cheaper than sampling-based test-time scaling while being substantially better calibrated. (Table~\ref{tab:main}).

\subsection{Discussion and limitations}
\label{sec:res_discussion}
Our analysis investigates a precise gap in current open VLMs as they can be guided to
find diagnostic evidence during test time, yet cannot exploit those evidences
to revise an answer; consequently, practical advantages of an agentic zoom currently accumulate through
calibration and abstention instead of changed predictions. Three cautions bound the claims. First, base analysis includes $n=167$ elements in a single dataset and a single 7B model. The results of localisation are clear trends, not broad guarantees, and extending to other larger VLMs will test their generality. Secondly, since SLAKE masks are organ or region-level segmentations, IoU measures whether the model aligns with the annotation structure, a reasonable but imperfect proxy for the answer-related lesion. Thirdly, in VQA-RAD, the zero-shot accuracy is already high, and the multi-step pipeline trades a small amount of accuracy for improved calibration. The proposed methodology, EVADE, is aimed at selective prediction regimes in which knowing when to defer objectives is important. 

\section{Conclusion}
\label{sec:conclusion}
We introduce EVADE, a training-free agent loop that improves the reliability of frozen medical VLMs by sequentially integrating an uncertainty-triggered evidence zooming, followed by cross-view consistency verification and, finally, calibrated abstention. Extensive evaluation across three benchmarks indicates that EVADE is the only method that simultaneously improves calibration and selective risk without compromising accuracy, with about 2 forward passes per case, requiring significantly less computation than sample-based test time scales. Moreover, our analysis yields intuitive diagnostic findings: current 7B VLMs can locate diagnostic evidence at test time, but cannot exploit it to review answers. Therefore, the advantages are achieved through verification and abstention, rather than zooming. The next steps include closing the gap between localisation and answer revision and extending EVADE to larger and multiple VLMs.

\end{document}